# ACIT: Attention-Guided Cross-Modal Interaction Transformer for Pedestrian Crossing Intention Prediction

Yuanzhe Li[1,*], *Graduate Student Member*, *IEEE*, and Steffen Müller[1]

*Abstract*— Predicting pedestrian crossing intention is crucial for autonomous vehicles to prevent pedestrian-related collisions. However, effectively extracting and integrating complementary cues from different types of data remains one of the major challenges. This paper proposes an attention-guided cross-modal interaction Transformer (ACIT) for pedestrian crossing intention prediction. ACIT leverages six visual and motion modalities, which are grouped into three interaction pairs: (1) Global semantic map and global optical flow, (2) Local RGB image and local optical flow, and (3) Ego-vehicle speed and pedestrian's bounding box. Within each visual interaction pair, a dual-path attention mechanism enhances salient regions within the primary modality through intra-modal self-attention and facilitates deep interactions with the auxiliary modality (i.e., optical flow) via optical flow-guided attention. Within the motion interaction pair, cross-modal attention is employed to model the cross-modal dynamics, enabling the effective extraction of complementary motion features. Beyond pairwise interactions, a multi-modal feature fusion module further facilitates cross-modal interactions at each time step. Furthermore, a Transformer-based temporal feature aggregation module is introduced to capture sequential dependencies. Experimental results demonstrate that ACIT outperforms state-of-the-art methods, achieving accuracy rates of 70% and 89% on the JAAD$_{beh}$ and JAAD$_{all}$ datasets, respectively. Extensive ablation studies are further conducted to investigate the contribution of different modules of ACIT.

## I. INTRODUCTION

The increasing deployment of autonomous vehicles (AVs) in urban environments has raised growing concerns regarding the safety of vulnerable road users (VRUs), particularly pedestrians. Predicting pedestrian crossing intention is an important function of AVs to improve the safety level, as it increases the available response time. However, accurately predicting the pedestrian's crossing intention remains challenging due to its diversity and the influence of multiple environmental factors, such as road layout, traffic dynamics, and interactions with other traffic participants [24].

Pedestrian crossing intention prediction has attracted growing research interest in recent years. Early studies primarily rely on single-modal data, such as pedestrian images [2] and pose keypoints [25]. However, the use of single-modal data alone is insufficient for reliable prediction, given the susceptibility of pedestrian intentions to multiple impacting factors. For instance, research employing only pedestrian images fails to capture the influence of other traffic participants on pedestrian behavior.

Recent studies have focused on fusing multimodal data to address this task. More visual modalities (e.g., semantic map [4], optical flow [5]) and motion modalities (e.g., bounding box [6], vehicle speed [7]) have been introduced as complementary inputs to enhance the understanding of pedestrian-specific and global contexts. In most studies, separate branches are designed for each modality to extract features from the entire observation sequence, followed by multimodal feature fusion for intention prediction. For instance, in [4], Global PCPA is proposed, where convolutional neural networks (CNNs) are employed to extract visual features from visual modalities, which are subsequently pooled and individually encoded together with data from other motion modalities by separate gated recurrent units (GRUs) for temporal modeling. A modality-level attention mechanism is finally applied to integrate the temporally fused features. In [9, 26], a similar fusion strategy is adopted, where Transformers are used to independently perform temporal feature fusion for different modalities, followed by multi-scale feature stacking or a Transformer-based modality attention for intention prediction.

However, two limitations remain to be addressed. (1) Most studies directly apply pooling operations without considering that the intermediate feature maps contain rich spatial information. Moreover, current methods typically rely on simple addition or concatenation of pooled features for visual feature fusion, ignoring the complementary spatial-scale information across visual modalities. (2) Most studies first perform temporal fusion within each modality and subsequently apply modality fusion. As a result, cross-modal interactions are confined to the temporally fused features, which ignores step-wise cross-modal interactions and fails to effectively capture the complementary cues.

In this paper, we propose an attention-guided cross-modal interaction Transformer (ACIT) for pedestrian crossing intention prediction. ACIT fuses features from six motion and visual modalities, which are organized into three interaction pairs based on their spatial coverage and physical properties. Our main contributions are summarized as follows: (1) A dual-path attention mechanism is proposed for visual interaction pairs, which enhances the intermediate feature map of the primary modality (i.e. local RGB image and global semantic map) via intra-modal self-attention and facilitates cross-modal interaction through optical flow-guided attention. (2) Dynamic interactions within the motion interaction pair are captured through a cross-modal attention mechanism. (3) A modality-temporal fusion order is adopted, where a multi-modal feature fusion module and a subsequent temporal feature aggregation module are designed to perform modality-level fusion beyond interaction pairs and to fuse features across the temporal dimension. (4) Extensive experiments on the JAAD dataset are conducted to validate the ACIT's performance.

[1]Chair of Automotive Engineering, Technische Universität Berlin, Berlin, 13355, Germany.
*Yuanzhe Li is the corresponding author of this paper. The E-mail is yuanzhe.li@campus.tu-berlin.de

## II. RELATED WORK

Early studies on pedestrian crossing intention prediction mainly focus on single-modal data. For instance, in [2], 2D CNNs are used to extract pedestrian appearance features from local image for the intention prediction, whereas in [22], 3D CNNs are applied to image sequence to extract temporally continuous pedestrian features. However, these methods overlook the complementary cues embedded in different modality streams. Recent studies have shown that multimodal feature fusion can enhance prediction accuracy by enriching contextual information and better modeling intrinsic and extrinsic influences on pedestrian behavior. A variety of multimodal fusion networks have been developed. In [4], Global PCPA is proposed, where features from visual modalities (local RGB images and global semantic maps) and motion modalities (vehicle speed, bounding boxes, and pose keypoints) are first temporally fused using GRUs, and the aggregated features are then fed into a modality attention module for the final prediction. In [5], optical flow, and in [23], distance information are additionally incorporated as complementary inputs to enhance the modeling of environmental cues relevant to pedestrian intentions.

The success of Transformers and their variants in handling sequential modeling and computer vision tasks has spurred the development of Transformer-based multimodal fusion networks. In [12], Action-ViT is proposed, where multiple frames of visual modalities are stacked into a single image and motion modalities are converted into pseudo-image formats, with a Vision Transformer (ViT) applied to extract spatiotemporal features. In [9], different Transformer blocks are employed to perform temporal fusion for different modalities, followed by a Transformer-based modality attention module to fuse the multimodal features.

## III. METHODOLOGY

### A. Problem Formulation

The task is to predict a pedestrian's crossing intention 1–2 seconds ahead based on a 0.5-second video (16 frames). The final frame is placed immediately before the event, defined as crossing initiation or the last observable frame in non-crossing cases. ACIT utilizes multimodal data, encompassing the following six modalities: (1) Global semantic map $G_S \in \mathbb{R}^{N \times 256 \times 256 \times 3}$, which encapsulates road structures and traffic information. $G_s$ is generated using pretrained SegFormer model [13]. (2) Global optical flow $G_{OF} \in \mathbb{R}^{N \times 256 \times 256 \times 3}$, which captures the relative motion between the environment and the ego-vehicle. $G_{OF}$ is generated using the pretrained RAFT model [15]. (3) Local RGB image $L_{RGB} \in \mathbb{R}^{N \times 256 \times 256 \times 3}$, which captures the pedestrian's appearance. $L_{RGB}$ is cropped to the original bounding box size. (4) Local optical flow $L_{OF} \in \mathbb{R}^{N \times 256 \times 256 \times 3}$, which captures the relative motion of the pedestrian with respect to the ego-vehicle. $L_{OF}$ is cropped from the $G_{OF}$ based on the bounding box. (5) Ego-vehicle's speed $S \in \mathbb{R}^{N \times 1}$, which represents the ego-vehicle's motion. $S$ is directly retrieved from the dataset annotations. (6) Pedestrian's bounding box $B \in \mathbb{R}^{N \times 4}$, which represents the pedestrian's relative position. $B$ is defined by the coordinates of the top-left and bottom-right corners. Each modality has a sequence of length $N=16$. The overall architecture of ACIT is illustrated in Fig. 1.

### B. Visual Feature Encoding

The visual feature encoding (VFE) module encodes the raw visual modality inputs into compact intermediate feature maps that preserve spatial information. In the VFE module, we adopt the pretrained Swin Transformer V2 Base model [16] as the backbone. Swin V2 combines CNN-like locality and hierarchical structures with Transformer-based long-range dependency modeling via non-overlapping local windows and cross-window connections. All visual modality data are resized to 256×256 to meet the input requirements of Swin V2. The intermediate feature maps from the final stage of Swin V2, denoted as $F_{L_{RGB}}, F_{L_{OF}}, F_{G_S}$, and $F_{G_{OF}} \in \mathbb{R}^{N \times 8 \times 8 \times 1024}$, captures compact visual cues from the corresponding modalities and are used for visual feature fusion.

### C. Attention-Guided Visual Modality Interaction

The attention-guided visual modality interaction (AVMI) module groups visual modalities into two interaction pairs based on spatial coverage: (1) Local RGB image $L_{RGB}$ and local optical flow $L_{OF}$, and (2) Global semantic map $G_S$ and global optical flow $G_{OF}$. The AVMI module takes $L_{RGB}$ and $G_S$ as primary modality, and $L_{OF}$ and $G_{OF}$ serve as auxiliary modality.

The core of AVMI module is a dual-path attention mechanism. Taking the interaction pair $L_{RGB}$ and $L_{OF}$ as an example, the architecture is illustrated in Fig. 2, which consists of an RGB self-attention (RGB-SA) path, and an optical flow-guided attention (OF-GA) path. At time step $\tau$, the feature maps $f^{\tau}_{L_{RGB}}$ and $f^{\tau}_{L_{OF}}$ are first processed by a 1×1 convolution layer with 256 filters to reduce the channel dimensions, thereby decreasing computational complexity. They are then flattened into token sequences of size 64×256, and positional encodings are added to preserve spatial information. In the RGB-SA path, intra-modal self-attention is applied to enhance critical spatial regions, represented by tokens within the RGB token sequence. While in the OF-GA path, optical flow-guided attention is utilized to strengthen cross-modal interaction between tokens from $L_{RGB}$ and $L_{OF}$. The process can be formulated as follows:

$$RGB-SA(Q_{RGB}, K_{RGB}, V_{RGB}) = \text{Softmax}\left(\frac{Q_{RGB} K_{RGB}^T}{\sqrt{d}}\right) V_{RGB} \quad (1)$$

$$OF-GA(Q'_{RGB}, K_{OF}, V_{OF}) = \text{Softmax}\left(\frac{Q'_{RGB} K_{OF}^T}{\sqrt{d}}\right) V_{OF} \quad (2)$$

Where $Q_{RGB}, K_{RGB}, V_{RGB}$ denote query, key and value generated from RGB tokens, respectively, $Q'_{RGB}$ also from RGB tokens and is used for cross-modal interaction, while $K_{OF}$ and $V_{OF}$ are obtained from optical flow tokens.

The resulting $RGB-SA(Q_{RGB}, K_{RGB}, V_{RGB})$ is added back to the token sequences of RGB images as residual components. To improve the adaptability of OF-GA path, a learnable weight parameter $\alpha$ is introduced to scale $OF\text{-}GA(Q'_{RGB}, K_{OF}, V_{OF})$, which is then added to the token sequence output of RGB-SA path through residual connections. $F_{G_S}$, and $F_{G_{OF}}$ are processed in a similar manner, where the token sequence obtained from each interaction pair is reshaped into feature maps of size $N \times 8 \times 8 \times 256$. The fused local RGB-optical flow features $F_L$ and the fused global semantic-optical flow features $F_G$ are obtained by applying global average pooling.

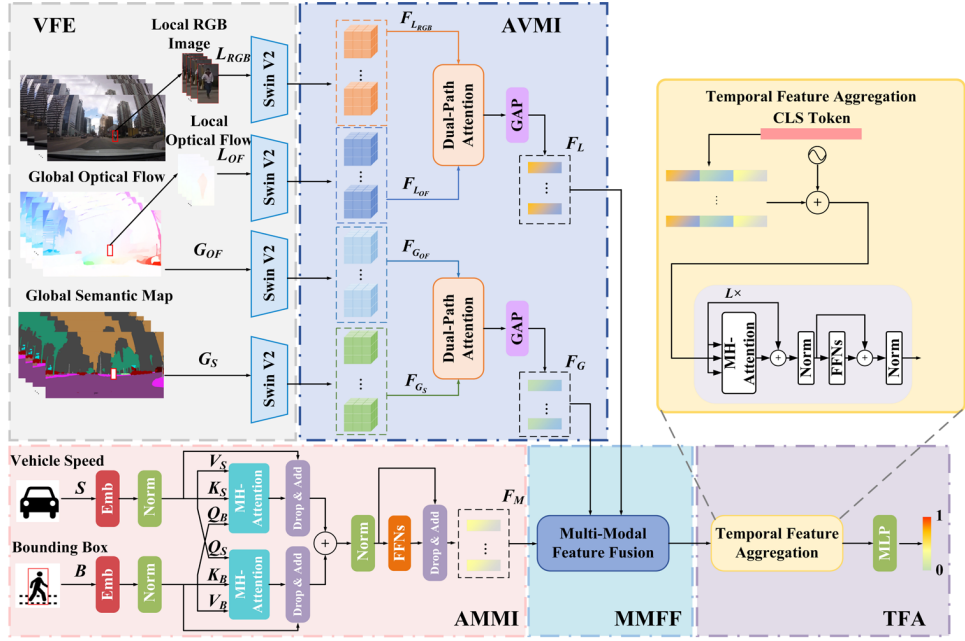

Fig. 1. The architecture of the proposed ACIT model.

### D. Attention-Guided Motion Modality Interaction

The attention-guided motion modality interaction (AMMI) module facilitates deep interactions between motion modalities and extracts complementary motion features. The original motion modality inputs $S \in \mathbb{R}^{N \times 1}$ and $B \in \mathbb{R}^{N \times 4}$ are first projected through individual embedding layers into high-dimensional representations of size $N \times 256$, followed by normalization, resulting in $X_S$ and $X_B$. Cross-modal attention is designed to model pairwise correlations between motion modalities, where two multi-head attention (MHA) units, each with $n$ parallel attention heads, are employed to concurrently focus on diverse representation subspaces. For modality $j$, the output of the MHA unit is given by:

$$\text{MHA}(Q_i, K_j, V_j) = [CA_1,...,CA_n]W_{CA}^O \quad (3)$$

$$CA_k(Q_{i,k}, K_{j,k}, V_{j,k}) = \text{Softmax}\left(\frac{Q_{i,k}K_{j,k}^T}{\sqrt{d}}\right)V_{j,k} \quad (4)$$

Where $i, j \in \{S, B\}$, $i \neq j$. $Q_{i,k}$ denotes the query of modality $i$ for attention head $k$, which is obtained by applying $W_{i,k}^Q$ to $X_i$. $K_{j,k}$ and $V_{j,k}$ are key and value of modality $j$ for attention head $k$, which are computed by applying the learnable weight matrix $W_{j,k}^K$ and $W_{j,k}^V$ to $X_j$.

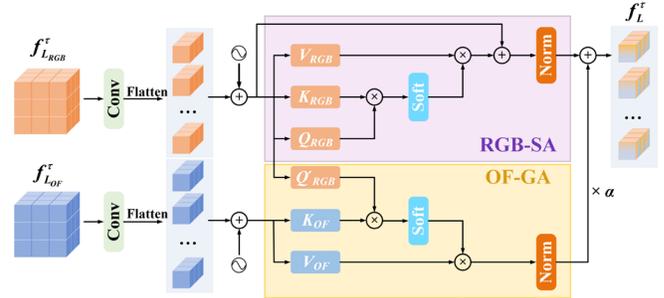

Fig. 2. The architecture of the dual-path attention.

The output of each MHA unit undergoes dropout and residual addition, after which the two branches are added. The result is then normalized and passed through a feed-forward network to enhance the motion features $F_M$ of size $N \times 256$.

### E. Multi-Modal Feature Fusion

The multi-modal feature fusion (MMFF) module employs inter-modal attention to capture fame-wise interactions among multimodal features. At each time step, cross-modal queries $Q_i^\tau$, keys $K_i^\tau$, and values $V_i^\tau$ are obtained by projecting the input features through weight matrices $W_i^Q$, $W_i^K$, $W_i^V$, where $i \in \{L, G, M\}$. Attention weights, indicating the correlation of modality $i$ to $j$, are computed as follows:

$$\alpha_{ij}^\tau = \frac{\exp\left(\frac{Q_i^\tau K_j^{\tau T}}{\sqrt{d}}\right)}{\sum_{k=1}^{3}\exp\left(\frac{Q_i^\tau K_k^{\tau T}}{\sqrt{d}}\right)} \quad (5)$$

The informative features from one modality can be propagated to others based on the learned correlations. The refined feature of modality *i* at time step $\tau$ is obtained:

$$\tilde{f}_i^\tau = \sum_{j=1}^{3} \alpha_{ij}^\tau V_j^\tau \quad (6)$$

Finally, the refined features of $F_L$, $F_G$, and $F_M$ are concatenated to form the fused feature sequence.

*F. Temporal Feature Aggregation*

Considering the varying importance across different time steps, a temporal feature aggregation (TFA) module is developed to model the intrinsic temporal dependencies. TFA module is implemented by a Transformer encoder [18]. A CLS token is incorporated into the fused feature sequence to construct an aggregated sequence representation, with positional encoding to preserve spatial information.

The Transformer encoder is configured with 2 layers and 6 heads, where each layer consists of a multi-head self-attention (MHSA) followed by a feed-forward network (FFN). The MHSA is composed of *m* parallel attention heads. For attention head *h*, the query $Q_h$, key $K_h$, and value $V_h$ are computed by projecting the input features through trainable matrices $W_h^Q$, $W_h^K$, $W_h^V$, respectively. Multiplying $V_h$ by the attention matrix allows each time step in the output to incorporate temporal information from the other steps, which can be formulated as

$$SA_h(Q_h, K_h, V_h) = \text{Softmax}\left(\frac{Q_h K_h^T}{\sqrt{d_h}}\right) V_h \quad (7)$$

The outputs of all self-attention heads are concatenated and projected through a trainable weight matrix, which can be expressed as:

$$\text{MHSA}(Q, K, V) = [SA_1, SA_2, ..., SA_m] W_{SA}^O \quad (8)$$

After applying residual connections and normalization, the output of the MHSA is fed into the FFN for feature enhancement. The CLS token output from the final layer of the TFA module is passed through a multilayer perceptron (MLP) layer to predict the pedestrian's intention.

## IV. EXPERIMENTS

*A. Datasets and Implementation Details*

ACIT is evaluated on the JAAD dataset [2], comprising on-board videos recorded with a monocular camera at 30 FPS. The JAAD dataset comprises two subsets: JAAD$_{beh}$, with 495 crossing and 191 non-crossing pedestrians, and JAAD$_{all}$, which further includes 2,100 visible pedestrians with non-crossing behavior. The dataset is divided into 177 videos for training, 117 for testing, and 29 for validation, following the benchmark setting in [6]. We extract 16-frame clips with a temporal overlap of 80% from the annotated sequences. ACIT is implemented using TensorFlow on RTX 3090 Ti GPU. We use dropout of 0.3, L2 regularization of 0.001 in the MLP layer, binary cross-entropy loss, and Adam optimizer with a learning rate of $2 \times 10^{-5}$. To address class imbalance, we applied class weights proportional to positive–negative sample ratios.

*B. Quantitative Results*

We compare ACIT with the state-of-the-art (SOTA) methods. The results are shown in Table I. Green and blue indicate the best and second-best results, respectively. Performance is evaluated using five metrics: accuracy (Acc), AUC, F1 score, precision, and recall, consistent with standard practice in related works.

On the JAAD$_{beh}$ dataset, ACIT achieves the highest accuracy and F1 score, reaching 70% and 79%, respectively, with an improvement of 1% over the second-best method, RU-LSTM [3]. ACIT also achieves the highest AUC and recall, reaching 62% and 93%, respectively, with the AUC comparable to RU-LSTM [3] and the recall exceeding TrouSPI-Net [14] by 2%. The precision of ACIT reaches 69%, ranking second and falling 2% behind V-PedCross [5].

On the JAAD$_{all}$ dataset, ACIT achieves the highest accuracy of 89%, comparable to STFF-MANet [1]. However, ACIT outperforms STFF-MANet [1] by 3%, 2%, and 7% in terms of AUC, F1 score, and recall, respectively. Moreover, ACIT achieves the highest AUC and F1 scores among all methods. Compared to Global PCPA [4], which ranks second in AUC, ACIT shows a 1% improvement. In terms of precision, ACIT ranks second, falling 3% behind STFF-MANet [1]. However, ACIT shows weaker recall, falling 9% behind PCPA [6].

The quantitative results demonstrate that the proposed network benefits from deep interactions across different modalities and effective fusion strategies, enabling more effective extraction and integration of complementary features and ultimately outperforming the SOTA methods.

*C. Ablation Study*

To assess the contribution of individual modules and the impact of the fusion order, we provide five architectural variants, as illustrated in Fig. 3. (1) ACIT-v1: The Transformer-based temporal feature fusion in the TFA module is replaced by a simple temporal pooling operation. (2) ACIT-v2: The inter-modal attention in MMFF module is substituted with a simple element-wise addition of multimodal features $F_L$, $F_G$, and $F_M$. (3) ACIT-v3: The cross-modal attention in the AMMI module is replaced by independent self-attention for each motion modality, followed by element-wise addition to fuse the motion features (4) ACIT-v4: The dual-path attention mechanism is replaced by performing independent average pooling on each feature map, followed by an addition operation to fuse the visual features. (5) ACIT- v5: A temporal-modality fusion order is adopted, where temporal fusion is first performed independently for each modality, followed by an inter-modal attention on the temporally fused features. The results of the ablation study are shown in Table II.

**Contribution of TFA module:** ACIT-v1 shows a decline across almost all evaluation metrics, with a notable decrease in accuracy, dropping by 3% and 2%, respectively. This is mainly attributed to the absence of temporal feature fusion, resulting in each time step being treated equally and without considering the varying contributions across the temporal dimension. In contrast, the TFA module of ACIT exploits the self-attention mechanism to effectively capture and model long-range dependencies within the temporal sequences.

TABLE I.  QUANTITATIVE RESULTS ON JAAD DATASET

| Models | JAAD_beh | | | | | JAAD_all | | | | |
|---|---|---|---|---|---|---|---|---|---|---|
| | Acc | AUC | F1 | Precision | Recall | Acc | AUC | F1 | Precision | Recall |
| SF-GRU [7] | 0.58 | 0.56 | 0.65 | 0.68 | 0.62 | 0.76 | 0.77 | 0.53 | 0.40 | 0.79 |
| FUSSI [19] | 0.59 | 0.58 | 0.69 | 0.66 | 0.73 | 0.60 | 0.72 | 0.40 | 0.27 | 0.73 |
| SingleRNN [20] | 0.60 | 0.54 | 0.70 | 0.65 | 0.76 | 0.78 | 0.77 | 0.54 | 0.42 | 0.75 |
| BiPed [21] | - | - | - | - | - | 0.83 | 0.79 | 0.60 | 0.52 | - |
| IntFormer [10] | 0.59 | 0.54 | 0.69 | - | - | 0.86 | 0.78 | 0.62 | - | - |
| PCPA [6] | 0.56 | 0.54 | 0.63 | 0.66 | 0.60 | 0.77 | 0.79 | 0.56 | 0.42 | **0.83** |
| Global PCPA [4] | 0.62 | 0.54 | 0.74 | 0.65 | 0.85 | 0.83 | **0.82** | 0.63 | 0.51 | **0.81** |
| MMH-PAP [17] | - | - | - | - | - | 0.84 | 0.80 | 0.62 | 0.54 | - |
| TrouSPI-Net [14] | 0.64 | 0.56 | 0.76 | 0.66 | **0.91** | 0.85 | 0.73 | 0.56 | 0.57 | 0.55 |
| ST CrossingPose [11] | 0.63 | 0.56 | 0.74 | 0.66 | 0.83 | - | - | - | - | - |
| V-PedCross [5] | 0.61 | 0.50 | 0.75 | **0.71** | 0.80 | 0.82 | 0.74 | 0.64 | 0.58 | 0.63 |
| RL [8] | 0.66 | 0.56 | 0.76 | - | - | - | - | - | - | - |
| RU-LSTM [3] | **0.69** | **0.62** | **0.78** | - | - | 0.86 | 0.78 | 0.62 | - | - |
| STFF-MANet [1] | 0.66 | 0.58 | 0.77 | 0.67 | 0.89 | **0.89** | 0.80 | **0.67** | **0.68** | 0.67 |
| ACIT | **0.70** | **0.62** | **0.79** | **0.69** | **0.93** | **0.89** | **0.83** | **0.69** | **0.65** | 0.74 |

**Contribution of MMFF module:** ACIT-v2 exhibits a similar trend, with accuracy drops of 4% and 3% compared to ACIT, respectively. This is because each modality is treated equally, which hinders the effective modeling of their different contributions. In contrast, the MMFF module of ACIT employs inter-modal attention to capture and model the complex interactions among different modalities at each time step, allowing the network to selectively focus on information that is crucial for accurate prediction.

**Contribution of AMMI module:** ACIT-v3 underperforms ACIT, with the most notable decreases observed in accuracy (3%/3%) and AUC (4%/6%). ACIT-v3 adopts independent feature extraction of motion modalities, ignoring the potential complementarity arising from interactions among different motion modalities. The cross-modal attention in the AMMI module effectively simulates the dynamic interaction process between the target pedestrian and the ego-vehicle, enabling better capture of dynamic cues.

**Contribution of AVMI module:** ACIT-v4 shows inferior performance compared to ACIT, with accuracy decreasing by 5% and 2%, respectively. This performance gap is attributed to ACIT-v4's insufficient utilization of the spatial information in the feature maps and its restricted interactions. In contrast, the dual-path attention mechanism in ACIT, which combines intra-modal self-attention with optical flow-guided attention, enables more effective capture of visual cues by fully leveraging spatial information.

**Impact of fusion order:** ACIT-v5 exhibits an overall performance decline compared to ACIT, with the accuracy decreasing by 3%. The performance drop is attributed to the temporal-modality fusion order adopted in ACIT-v5, which overlooks the potential of multimodal interactions at time-step level. In ACIT-v5, multimodal interactions occur only after the temporal features have been fused. In contrast, ACIT adopts a modality-temporal fusion order, enabling multi-modal interactions at each time step and fully exploiti-ng complementary information across modalities.

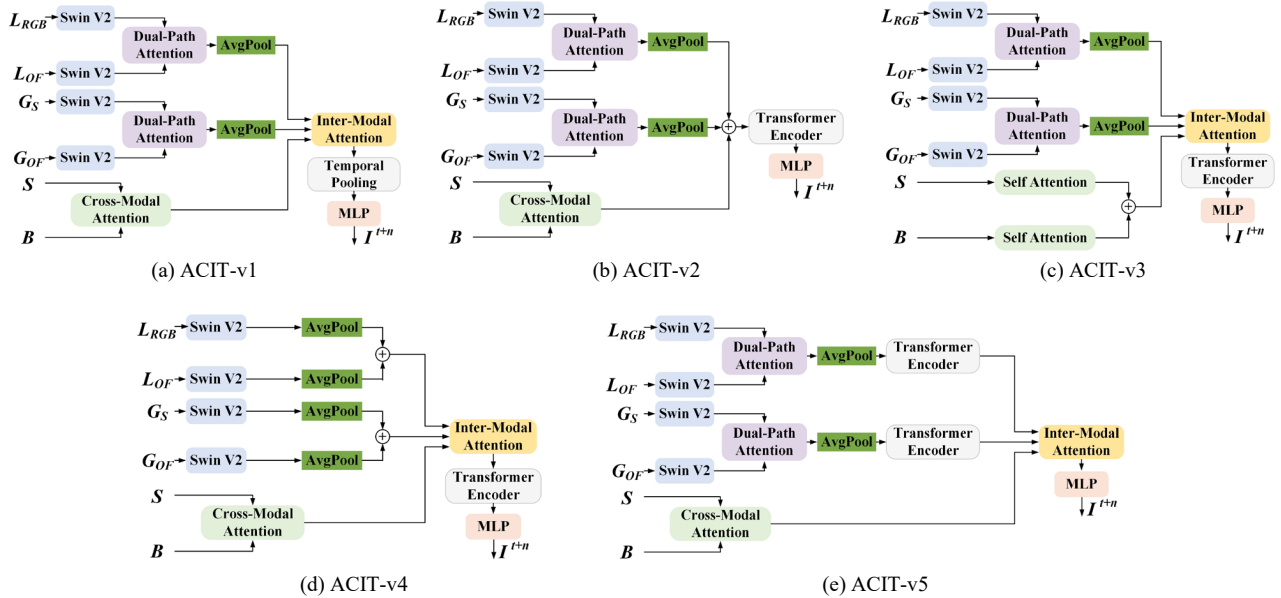

Fig. 3. Architectural illustration of ACIT variants.

TABLE II. ABLATION STUDY ON JAAD DATASET

| Variant | JAAD_beh/JAAD_all | | | | |
|---|---|---|---|---|---|
| | Acc | AUC | F1 | Precision | Recall |
| ACIT-v1 | 0.67/0.87 | 0.58/0.81 | 0.78/0.66 | 0.67/0.61 | **0.93**/0.71 |
| ACIT-v2 | 0.66/0.86 | 0.58/0.81 | 0.77/0.65 | 0.67/0.58 | 0.91/0.73 |
| ACIT-v3 | 0.67/0.86 | 0.58/0.77 | 0.78/0.61 | 0.67/0.59 | 0.92/0.64 |
| ACIT-v4 | 0.65/0.87 | 0.56/0.78 | 0.77/0.63 | 0.66/0.61 | **0.93**/0.66 |
| ACIT-v5 | 0.67/0.86 | 0.59/0.80 | 0.78/0.64 | 0.67/0.59 | 0.92/0.69 |
| ACIT | **0.70/0.89** | **0.62/0.83** | **0.79/0.69** | **0.69/0.65** | **0.93/0.74** |

*D. Qualitative Results*

Fig. 4 presents prediction results in several typical cases, where we primarily compare ACIT with Global PCPA (GP) [4]. The target pedestrian is framed out by a red bounding box. GT represents the ground truth, while C and NC denote crossing and no crossing, respectively. It shows that ACIT can effectively predict pedestrians' intentions in the following cases: In case (a): An adult is walking along one side of a parking lot while approaching. In case (b): A pedestrian slowly moves from the roadside toward the street. In case (c): In the distance, a pedestrian is standing at the roadside, looking toward the ego-vehicle and hesitating to cross. In case (d): A pedestrian changes direction at the intersection and hesitantly tries to cross the street. It can be observed that ACIT consistently correctly predicts the pedestrian's crossing intention, while Global PCPA has difficulties. The competitive performance of ACIT can be attributed to the effective deep interactions between modalities and the modality-temporal fusion strategy, which jointly enhances the model's ability to capture subtle environmental and situational cues, thereby improving the performance of pedestrian intention prediction.

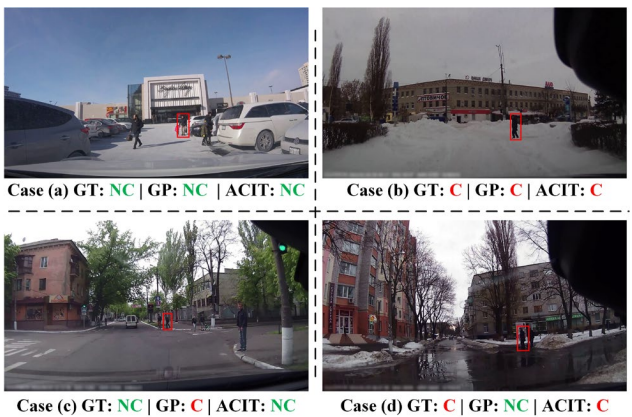

Fig. 4. Qualitative results of Global PCPA and ACIT on JAAD dataset.

*E. Computational Cost Analysis*

We evaluate the real-time performance and computational cost of ACIT by comparing it with several baseline methods, including PCPA [20], Global PCPA [21], and FUSSI [47], all of which integrate both visual and non-visual modalities. The results are shown in Table III. Green and blue indicate the best and second-best results, respectively. Although ACIT integrates the most modalities among all methods, it maintains a relatively compact architecture, with a model size of 62.50 MB and 5.15 million parameters, ranking second smallest overall. ACIT has a slightly slower inference time (43.93 ms) than PCPA [6] and FUSSI [19], but achieves substantially higher accuracy: +14% and +11% on JAAD_beh dataset, and +12% and +29% on JAAD_all dataset. The results of the computational cost analysis demonstrate that ACIT holds significant potential for real-time deployment.

TABLE III. COMPARISON OF COMPUTATIONAL COST

| Models | Size (MB) | Parameters (Million) | Inference Time (ms) |
|---|---|---|---|
| PCPA [6] | 118.80 | 31.17 | **38.60** |
| Global PCPA [4] | 374.20 | 60.92 | 70.83 |
| FUSSI [19] | **8.40** | **1.00** | **34.92** |
| ACIT | **62.50** | **5.15** | 43.93 |

V. CONCLUSIONS

In this paper, we propose a novel attention-guided cross-modal interaction Transformer for pedestrian crossing intention prediction. The proposed network groups multiple modalities into three interaction pairs based on the spatial coverage and physical properties. For each visual interaction pair, a dual-path attention mechanism is proposed to facilitate both intra-modal and cross-modal interactions through various attention mechanisms. In the motion interaction pair, cross-modal attention is employed to capture complementary motion features. In addition, multi-modal feature fusion module and temporal feature aggregation module are introduced to enable modality attention after pairwise interactions and to aggregate temporal features. Experimental results show that the proposed network outperforms state-of-the-art methods. Future work will focus on model compression and real-world deployment.